\documentclass{article}

\usepackage{microtype}
\usepackage{graphicx}
\usepackage{subfigure}
\usepackage{booktabs} 

\usepackage{hyperref}

\usepackage[nonarchival]{icml2023}

\usepackage{amsmath}
\usepackage{amssymb}
\usepackage{mathtools}
\usepackage{amsthm}

\usepackage{amsmath,amsfonts,bm}

\def\eqref#1{equation~\ref{#1}}

\def\1{\bm{1}}

\DeclareMathAlphabet{\mathsfit}{\encodingdefault}{\sfdefault}{m}{sl}
\SetMathAlphabet{\mathsfit}{bold}{\encodingdefault}{\sfdefault}{bx}{n}

\newcommand{\E}{\mathbb{E}}

\newcommand{\R}{\mathbb{R}}

\DeclareMathOperator*{\argmin}{arg\,min}

\usepackage{mathabx}
\usepackage{booktabs}
\usepackage{arydshln}
\usepackage{dsfont}
\usepackage{hyperref}
\usepackage{url}
\usepackage{scalerel,amssymb}

\usepackage{bm,amsmath}
\usepackage{tikz}
\usetikzlibrary{positioning, calc}

\renewcommand{\H}{\mathbb{H}}
\renewcommand{\S}{\mathbb{S}}

\renewcommand{\d}{\mathrm{d}}

\DeclareMathOperator{\diam}{diam}

\renewcommand{\dh}{\mathrm{d}_{\mathrm{H}}}
\newcommand{\dhg}{\mathrm{d}_{\mathrm{GH}}}
\newcommand{\ES}{\mathrm{ES}}
\newcommand{\be}{B_{\E^2}}
\newcommand{\bs}{B_{\S^2}}
\newcommand{\bh}{B_{\H^2}}

\DeclareMathOperator{\id}{id}

\renewcommand{\geq}{\geqslant}
\renewcommand{\leq}{\leqslant}

\renewcommand{\epsilon}{\varepsilon}

\newcommand{\rarrow}{\rightarrow}

\renewcommand{\d}{\mathrm{d}}

\newcommand{\en}{\mathbb{E}^n}
\newcommand{\sn}{\mathbb{S}^n}
\newcommand{\hn}{\mathbb{H}^n}

\newcommand{\hb}{B^1_{\mathbb{H}^n}}

\usepackage[capitalize,noabbrev]{cleveref}

\theoremstyle{plain}

\theoremstyle{definition}

\theoremstyle{remark}

\usepackage[textsize=tiny]{todonotes}

\icmltitlerunning{Gromov-Hausdorff Distances for Comparing Product Manifolds of Model Spaces}

\begin{document}

\twocolumn[
\icmltitle{Gromov-Hausdorff Distances \\ for Comparing Product Manifolds of Model Spaces}




\begin{icmlauthorlist}
\icmlauthor{Haitz Sáez de Ocáriz Borde}{ORI}
\icmlauthor{Álvaro Arroyo}{OMI}
\icmlauthor{Ismael Morales López}{math}
\icmlauthor{Ingmar Posner}{ORI}
\icmlauthor{Xiaowen Dong}{OMI,MLRG}
\end{icmlauthorlist}

\icmlaffiliation{ORI}{Oxford Robotics Institute, University of Oxford}
\icmlaffiliation{OMI}{Oxford-Man Institute, University of Oxford}
\icmlaffiliation{math}{Mathematical Institute, University of Oxford}
\icmlaffiliation{MLRG}{Machine Learning Research Group, University of Oxford}

\icmlcorrespondingauthor{Haitz Sáez de Ocáriz Borde}{haitz@oxfordrobotics.institute}

\icmlkeywords{Machine Learning, ICML}

\vskip 0.3in
]



\printAffiliationsAndNotice{}  

\begin{abstract}
Recent studies propose enhancing machine learning models by aligning the geometric characteristics of the latent space with the underlying data structure. Instead of relying solely on Euclidean space, researchers have suggested using hyperbolic and spherical spaces with constant curvature, or their combinations (known as \textit{product manifolds}), to improve model performance. However, there exists no principled technique to determine the best latent product manifold signature, which refers to the choice and dimensionality of manifold components. To address this, we introduce a novel notion of distance between candidate latent geometries using the Gromov-Hausdorff distance from metric geometry. We propose using a graph search space that uses the estimated Gromov-Hausdorff distances to search for the optimal latent geometry. In this work we focus on providing a description of an algorithm to compute the Gromov-Hausdorff distance between model spaces and its computational implementation.
\end{abstract}

\section{Introduction}
\label{submission}

Recent research has shown a growing interest in using concepts from differential geometry and topology to improve learning algorithms~\citep{Bortoli2022RiemannianSG,Hensel2021ASO,Chamberlain2021,Huang2022RiemannianDM,barbero2022sheaf,barbero2022sheaf2}. While Euclidean spaces have been commonly used in machine learning, it has been found that using geometries that align better with the underlying data structure can lead to significant improvements. One approach is to use constant curvature model spaces like the Poincaré ball~\citep{Mathieu2019}, hyperboloid~\citep{HGCN}, or hypersphere~\citep{Zhao2019LatentVO} to encode latent representations of data. Another approach is to use product spaces~\citep{Gu2018,Shopek2019,Ocariz2022,Projections2023,Zhang2020ProductML}, which combine multiple model spaces, allowing for more complex representations while maintaining computational tractability. However, determining the optimal configuration of product manifolds for representing data is currently done heuristically, lacking efficiency and a principled framework. In this work, we propose a computational approximation of the Gromov-Hausdorff distances between product manifolds of model spaces. This measure of closeness between manifolds can be useful for machine learning applications where finding an appropriate manifold to represent the embedding space of data or latent representations is crucial.
\vspace{-4mm}
\section{Background}
\label{sec:Background}
\vspace{-1mm}

\textbf{Differential Geometry and Product Manifolds.} Constant curvature model spaces refer to Riemannian manifolds that have a consistent sectional curvature, meaning the curvature remains the same in all directions within a 2D plane (and can be extended to higher dimensions)~\cite{book_gallier}. Examples of such spaces include Euclidean space, hyperbolic space, and the sphere. Euclidean space has zero curvature, hyperbolic space has negative curvature, and the sphere has positive curvature. Constant curvature model spaces are commonly used as reference spaces for comparative analysis and machine learning tasks involving non-Euclidean data. A product manifold can be constructed by taking the Cartesian product $\mathcal{P} = \bigtimes_{i=1}^{n_{\mathcal{P}}} \mathcal{M}_{K{i}}^{d_i}$ of $n_{\mathcal{P}}$ manifolds with curvature $K_{i}$ and dimensionality $d_i$. It is important to note that $n_{\mathcal{P}}$ and $d_{i}$ are hyperparameters that define the product manifold $\mathcal{P}$ and need to be predetermined. On the other hand, the curvature of each model space $K_i$ can be learned using gradient descent. It is worth mentioning that while the product manifold construction allows for the generation of more complex embedding spaces than the original constant curvature model spaces, it does not enable the generation of arbitrary manifolds or provide control over local curvature. In this study, we are interested in geometries that can be represented employing model space Riemannian manifolds and Cartesian products of such manifolds.

\textbf{Hausdorff and Gromov-Hausdorff distances.} The Hausdorff distance~\citep{Jungeblut2021TheCO} between two subsets, $A$ and $B$, in a metric space X with the metric $d_X$, can be defined as follows: $\dh^X(A, B)=\max \left(\sup_{a\in A} d_X(a, B), \, \sup_{b\in B} d_X(b, A)\right).$ This quantity is infinite, therefore we will restrict our calculations to compact subsets $A$ and $B$. In this particular instance, we define  $\dh(A, B)$ as the smallest real number $c\geq 0$ such that for every $a\in A$ and every $b\in B$ there exist $a'\in A$ and $b'\in B$ such that both $d_X(a, b')$ and $d_X(a', b)$ are at most $c$. It is worth mentioning that the previous definition does not impose the need for differentiable structures on $X$, $A$, and $B$. They can simply be metric spaces. This enables us to tell apart the metric characteristics of Euclidean, hyperbolic, and spherical geometries beyond analytical concepts like curvature. Nevertheless, the aforementioned definition restricts the comparison of spaces $A$ and $B$ to instances where they are embedded in a particular metric space $X$. To assess the notion of distance between them, we will employ the Gromov-Hausdorff distance. Given a metric space $X$ and two isometric embeddings $f: A\rarrow X$ and $g: B\rarrow X$, we define $\dh^{X, f, g} (A, B)=\dh^X (f(A), g(B)).$ Next, considering two metric spaces $A$ and $B$, we denote by $\ES (A, B)$, the ``embedding spaces of $A$ and $B$'', as the triple $(X, f, g)$ where $X$ is a metric space and $f: A\rarrow X$ and $g: B\rarrow X$ are isometric embeddings. The \textit{Gromov-Hausdorff distance between $A$ and $B$} is defined as: $\dhg (A, B)=\inf_{(X, f, g)\in \ES(A, B)} \dh^{X, f, g} (f(A), g(B)).$

\section{Methodology: Gromov-Hausdorff Distances for Comparing Product Manifolds}

The Hausdorff distance is a measure of dissimilarity between two sets of points, in our case representing discretized versions of continuous manifolds. An algorithm proposed by~\citep{Taha2015AnEA} efficiently computes the exact Hausdorff distance between point sets but assumes that the point sets reside in the same space and have the same dimensionality, which limits its applicability. Gromov-Hausdorff distances allow for measuring the distance between metric spaces that are not inherently embedded in a common ambient space. In the application described, distances need to be calculated between Euclidean space, spherical space, and hyperbolic space, as well as products of these. This work focuses on describing an algorithm for obtaining computational upper bounds for the distances between Euclidean space and hyperbolic space, as well as between spherical space and hyperbolic space. The Gromov-Hausdorff distance between Euclidean space and spherical space can be estimated analytically and it is later provided in Table~\ref{tab:gh_coeffs}, Section~\ref{Computational Estimation of the Gromov-Hausdorff Distances}. These quantities can also be used to infer Gromov-Hausdorff distances between product manifolds.


\textbf{Strategy.} There exist multiple isometric embeddings of spaces $\mathbb{E}^n$ and $\mathbb{S}^n$ into $\mathbb{E}^{6n-6}$. While these first two spaces can already be embedded in $\mathbb{E}^{n+1}$, considering this higher-dimensional Euclidean space is beneficial because it allows for the isometric embedding of $\mathbb{H}^n$. This common underlying space enables the computation of Hausdorff distances between the geometries $\mathbb{E}^n$, $\mathbb{S}^n$, and $\mathbb{H}^n$, ultimately leading to the estimation of their mutual Gromov-Hausdorff distance. The embedding of $\mathbb{H}^n$ into $\mathbb{E}^{6n-6}$, which we describe, can be found in~\citet{Wol87} and is made explicit in~\citet{Bla55}. For a broader context and related advancements in the isometric embedding of homogeneous spaces into higher-dimensional ones, we recommend referring to Chapter 5 of the book by~\citet{Bra03}. In our experiments, we will primarily focus on working with product manifolds generated based on constant curvature model spaces of dimension $n=2$, and we will denote the embedding of $\mathbb{H}^2$ as $F: \mathbb{H}^2 \rightarrow \mathbb{E}^6$. Now we can summarise our strategy to estimate $\dhg(\be, \bh)$ as follows (for $\dhg(\be, \bh)$ it will be entirely analogous). The first step consists of approximating our infinite smooth spaces by finite discrete ones. For this,   we consider several collections of points $\{P_i\}_{i\in I}$ in $\E^2$ that are sufficiently well distributed. The exponential map can be applied to the collection of points ${\exp: T_0\H^2\cong \R^2\rarrow \bh}$ to get several collections of points $Q$ in $\bh$ (again, well distributed by construction). In addition, we will consider several isometric embeddings $f_k: \be\rarrow \R^6$. Hence, we take 
\begin{align}
    \dhg(\be, \bh) &\approx\min_{i, j, k} \dh^{\R^6, f_k, F}(P_i, Q_j) \nonumber \\
    &=\min_{i, j, k} \dh^{\R^6}(f_k(P_i), F(Q_j)).
    \label{GH_eq}
\end{align}
Note that here we restrict ourselves to $\mathbb{E}^2$, $\mathbb{S}^2$, and $\mathbb{H}^2$ for computational simplicity, but the approach is generalizable to $\mathbb{E}^n$, $\mathbb{S}^n$, and $\mathbb{H}^n$.

\textbf{Model Space Manifold Discretization.} To approximate the mutual Gromov-Hausdorff distance between the smooth spaces $\mathbb{E}^2$ and $\mathbb{H}^2$, we adopt a discretized representation of these spaces. We achieve this by generating well-distributed sets of points in $\mathbb{E}^2$ and $\mathbb{H}^2$ through the application of the exponential map to a collection of points in $\mathbb{R}^2$. Additionally, we consider multiple isometric embeddings of $\mathbb{E}^2$ into $\mathbb{R}^6$. The estimation of the Gromov-Hausdorff distance involves computing the minimum Hausdorff distance between the point sets derived from the collections in $\mathbb{R}^6$ and the isometric embedding of $\mathbb{H}^2$ into $\mathbb{R}^{6}$. The same approach applies when comparing spherical space. All our computations will be done in dimension two, so we will simply precise how to generate points in $\be$, $\bs$ and $\bh$. For $\be$ and $\bs$, we will use very elementary trigonometry. For $\bh$, we will need an explicit description of the exponential map of $\H^2$. Using the descriptions $\be=\{(r\cos (t), r \sin (t)): r\in [0, 1], t\in [0, 2\pi)\},$ and $\bs=\{(\sin(\beta)\cos(\alpha), \sin(\beta)\sin(\alpha), \cos(\beta)): \alpha\in [0, 2\pi), \beta\in [0, 1]\}$. This way, it will be easy to generate collections of points in $\be$ and $\bs$. As we anticipated above in the outline of our strategy to estimate $\dhg(\be, \bh)$, in order to give explicit  well-distributed collection of points in $\bh$, it is enough to give a well-distributed collection of points in $\be$ and consider its image under the exponential map $\exp_0: \be\rarrow \bh$, where we have identified $\be$ with the ball of radius one of $\R^2\cong T_0\H^2$, i.e. the tangent space of $\H^2$ at the point $0$.

\textbf{Mapping function to embed $\H^n$ into $\E^{6n-6}$.} We want to define an isometric embedding of $\hb$ into $\E^{6n-6}$, that is, $F$. This higher dimensional space is     our candidate to fit in the three geometries $\en$, $\hn$ and $\sn$ to compute their Hausdorff distances as subspaces of $\E^{6n-6}$ and hence estimate their Gromov-Hausdorff distances. Before describing such embedding, we introduce several preliminary auxiliary functions. Let $\chi(t)=\sin (\pi t)\cdot e^{-\sin^{-2}(\pi t)}$ for non-integer values of $t$. A priori, the inverse of $\sin(0)=0$ does not make sense but since $\lim_{t\rightarrow 0^+} \chi (t)=\lim_{t\rightarrow 0^-} \chi (t)=0$, we can set $\chi(0)=0$ so it is still continuous. In fact, it is smooth and, in particular, integrable. We can say the same at all points when $t$ is an integer, so we set $\chi(t)=0$ for all integers $t$ and we obtain an smooth function $\chi$ defined on $\R$: $A = \int_0^1 \chi(t) dt.$ We also define
\begin{align}
	\psi_1(x)=\sqrt{\frac{1}{A}\cdot \int_0^{1+x}\chi(t)\, \d t}
\end{align}
\vspace{-2mm}
\begin{align}
	\psi_2(x)=\sqrt{\frac{1}{A}\cdot \int_0^{x}\chi(t)\, \d t}
\end{align}
 We set $c$ to be the constant $c=2\max \left\{G_1, G_2\right\}$ defined in terms of
 \begin{align}
	 G_i=\left\|\frac{\d }{\d x}\left(\sinh(x)\cdot \psi_i(x)\right)\right\|_{L^{\infty}[-2, 2]},\ i=\{1,2\}
 \end{align}
Further, we make use of two auxiliary functions 
\begin{align}
	h(x, y)=\frac{\sinh(x)}{c}\Big(&  \psi_1(x)\cos (c\cdot y), \psi_1(x)\sin (c\cdot y),  \nonumber  \\
	& \psi_2(x)\cos (c\cdot y), \psi_2(x)\sin (c\cdot y) \Big)
\end{align}
\begin{align}
	\psi(x, y)=\left(\sinh^{-1}(y e^x), \log (\sqrt{e^{-2x}+y^2})\right).
\end{align}
which allows us to define 
\begin{align}
\footnotesize
 f_0(x, y)=\Big(& \int_0^{\sinh^{-1}(ye^x)}\sqrt{1-\epsilon(t)^2 } \d t,  \nonumber \\
  &\log (\sqrt{e^{-2x}+y^2}),h(\psi(x, y))\Big), 
\end{align}
with $\epsilon$ being $\epsilon = \frac{G_1^2+G_2^2}{c^2}.$ This way, we can set 
\begin{align}
    &f(x, y_1, \dots, y_{n-1}) =  \nonumber \\
    &=   \kappa\big( f_0(x, \sqrt{n-1} y_1), \dots, f_0(x, \sqrt{n-1}y_{n-1} \big)
\end{align}
with $\kappa=\frac{1}{\sqrt{n-1}}$. Recall that in our case, we use the function $F(\cdot) = f(n=2,\cdot)$ to map the set of points $Q$ in $\mathbb{H}^2$ to $\mathbb{R}^6$, and for computing
\vspace{-2mm}
\begin{align}
    \dhg(\be, \bh)&\approx\min_{i, j, k} \dh^{\R^6, f_k, F}(P_i, Q_j)  \nonumber\\
   & =\min_{i, j, k} \dh^{\R^6}(f_k(P_i), F(Q_j)).
\end{align}
\section{Computational Estimation of the Gromov-Hausdorff Distances}
\label{Computational Estimation of the Gromov-Hausdorff Distances}
Given that the Gromov-Hausdorff distance between Euclidean and Spherical Space can be calculated analytically, we will focus on the numerical algorithm required to compute the distances between $\mathbb{E}^n$ and $\mathbb{H}^n$, and $\mathbb{S}^n$ and $\mathbb{H}^n$.

\textbf{Computational Implementation of the Gromov-Hausdorff Distance between the Remaining Model Spaces of Constant Curvature.} In this section, we explore the process of discretizing the model spaces to generate points within balls of radius one, considering practical considerations. The Euclidean, hyperbolic, and spherical spaces are continuous manifolds, and in order to approximate the Gromov-Hausdorff distance between them, we need to discretize these spaces. For the Euclidean plane, we sample $\be=\{(r\cos (t), r \sin (t)): r\in [0, 1], t\in [0, 2\pi)\}$, with a discretization of $10,000$ points in both $r$ and $t$. In order to generate points in $\bh$ (hyperbolic space), we will use the points in $\be$ (Euclidean space) as a reference. We will employ the exponential map and coordinate transformations to convert the points in $\be$ into points in $\bh$. To avoid numerical instabilities, we restrict the points to $\be^{\prime}=\{(r\cos (t), r \sin (t)): r\in [0.00000001, 0.97], t\in [0, 2\pi)\}$ and use a discretization of $10,000$. To sample points for the spherical space, $\bs=\{(\sin(\beta)\cos(\alpha), \sin(\beta)\sin(\alpha), \cos(\beta)): \alpha\in [0, 2\pi), \beta\in [0, 1]\},$ we use a discretization of $100$ for both $\alpha$ and $\beta$. The level of discretization was selected as a balance between resolution and computational time. We noticed that the Gromov-Hausdorff distance reached a stable state with our chosen discretization. Nevertheless, due to the inherent characteristics of the Gromov-Hausdorff distance, it is challenging to determine whether increased discretization might reveal any unexpected behavior.

\textbf{Optimizing the Embedding Functions for the Euclidean and Spherical Spaces.} Next to approximate the Gromov-Hausdorff distances:
\begin{align}
    \dhg(\be, \bh)&\approx\min_{i, j, k} \dh^{\R^6, f_k, F}(P_i^{H}, Q_j) \nonumber\\
    &=\min_{i, j, k} \dh^{\R^6}(f_k(P_i^{H}), F(Q_j))\ ,
\end{align}
\vspace{-7mm}
\begin{align}
    \dhg(\bs, \bh)&\approx\min_{i, j, k} \dh^{\R^6, g_k, F}(P_i^{S}, Q_j) \nonumber\\
    &=\min_{i, j, k} \dh^{\R^6}(g_k(P_i^{S}), F(Q_j))\ ,
    \vspace{-6mm}
\end{align}
we employ the following functions to embed $\be$ (Euclidean space) and $\bs$ (spherical space) into $\mathbb{E}^{6n-6}$. To optimize for $f_k$, we explore all possible permutations of the basis vectors of $\mathbb{E}^6$: ${e_1,e_2,e_3,e_4,e_5,e_6}$. For each $f_k$, we consider two elements ${e_i,e_j}$ and utilize those dimensions to embed $\mathbb{E}^2$ within $\mathbb{E}^6$. In theory, we should also take into account a small offset represented by $F(\mathbf{0})=\mathbf{0}$, but in this case, the offset is zero. Moreover, during the optimization process, we incorporate a small vector (with all entries, except one dimension, set to zero) into the mapping function. This vector allows us to translate the plane in different directions, with an offset ranging between $-0.5$ and $0.5$. We consider a total of 100 steps within this range. To optimize for $g_k$, we follow a similar procedure where we explore all permutations of the basis vectors. However, in this case, we consider three basis vectors instead of two. For each permutation family $P_i^S$, we not only consider the original permutation but also its negative counterpart $-P_i^S$. We also offset the mapping function as described before. It is important to note that, in practice, $\be$ and $\bs$ are discretized as $P_i^{H}$ and $P_i^{S}$, respectively.

\textbf{Results.} The estimated Gromov-Haussdorff distances are presented in Table 1.

\begin{table}[H]

\caption{Estimated Gromov-Hausdorff distances (up to two decimal  places) between model spaces and corresponding edge weights in the graph search space.}
\centering
\begin{tabular}{lcc}
\toprule
Comparison Pair & $d_{\text{GH}}(\cdot)$ &  $w_{(\cdot)}$ \\\midrule
$(\mathbb{E}^2, \mathbb{S}^2)$   &   0.23 & 4.35\\
$(\mathbb{E}^2, \mathbb{H}^2)$   & 0.77 & 1.30 \\
$(\mathbb{S}^2, \mathbb{H}^2)$    &   0.84 & 1.20\\
\bottomrule   
\label{tab:gh_coeffs}
\end{tabular}

\end{table}

\section{Applications: Graph Search Space}

As outlined in previous sections of this paper, we utilize Cartesian products of model spaces with constant curvature to generate potential latent geometries. Our current emphasis is on constructing a search space specifically tailored for identifying the optimal latent geometry, where \textit{optimal} denotes the latent geometry that exhibits the highest performance for a given downstream task. The search space is represented as a graph, enabling the utilization of established search algorithms. In this setting, we focus on a scenario where the function $\mathfrak{f}(\cdot)$ to be minimized is defined on the graph nodes. The objective of employing a search algorithm is to locate the node corresponding to the minimum value, denoted as $v^{*}=\argmin_{v\in\mathcal{V}} \mathfrak{f}(v)$. In our configuration, each node within the graph corresponds to a distinct latent product manifold, and the value assigned to the node represents the validation set performance of a neural network architecture employing that particular latent geometry. To determine edge weights, we utilize the inverse of the Gromov-Hausdorff distance between product manifolds. We denote the edge weight between model spaces $\mathcal{M}_1$ and $\mathcal{M}_2$ as $w_{\mathcal{M}_1, \mathcal{M}_2}$, representing the inverse Gromov-Hausdorff distance. Edge weights are reported in Table \ref{tab:gh_coeffs}. Further, the Gromov-Hausdorff distance between manifolds of different dimensions is one, see Appendix~\ref{Upper Bound}. 

To provide additional structure, we suggest imposing connections between only nodes corresponding to product manifolds that differ by one model space. For instance, $\mathbb{E}^2\times\mathbb{H}^2$ and $\mathbb{E}^2\times\mathbb{S}^2$ would be connected with edge weighting $w_{\mathbb{H}^2, \mathbb{S}^2}$  while $\mathbb{S}^2\times\mathbb{S}^2$ and $\mathbb{E}^2\times\mathbb{E}^2$ would have no connection in the graph. Moreover, the connectivity between product manifolds of different dimensions adheres to a consistent principle. For example, there would exist a connection with a strength of one between $\mathbb{E}^2\times\mathbb{H}^2$ and $\mathbb{E}^2\times\mathbb{H}^2\times\mathbb{S}^2$. However, there would be no connection between $\mathbb{E}^2$ and $\mathbb{E}^2\times\mathbb{E}^2\times\mathbb{E}^2$, or between $\mathbb{H}^2$ and $\mathbb{E}^2\times\mathbb{S}^2$. This construction introduces a sense of directionality to the graph and results in clusters of product manifolds sharing the same dimension. Additionally, it is worth noting that in practice, there are only four distinct edge weights utilized. As an illustration, the connectivity strength between $\mathbb{E}^2\times\mathbb{H}^2$ and $\mathbb{S}^2\times\mathbb{H}^2$ is $w_{\mathbb{E}^2, \mathbb{S}^2}$  since $\dhg(\mathbb{E}^2\times\mathbb{H}^2, \mathbb{S}^2\times\mathbb{H}^2) = \dhg(\mathbb{E}^2, \mathbb{S}^2) + \dhg(\mathbb{H}^2, \mathbb{H}^2) = \dhg(\mathbb{E}^2, \mathbb{S}^2)$ given that $\dhg(\mathbb{H}^2, \mathbb{H}^2) = 0$. Visual representations of the graph search space can be found in Appendix \ref{appendix:fig}.

\section{Conclusion}
We have described an algorithm for estimating the Gromov-Hausdorff distance between model spaces and product manifolds of these. Our computational approximations of these distances can be leveraged to compare and provide a mathematically grounded sense of \textit{closeness} between possible latent geometry candidates for machine learning applications. In particular, we suggest that a Gromov-Hausdorff-informed graph search space can be constructed and combined with search algorithms.

\section{Acknowledgements}

AA thanks the Rafael del Pino Foundation for financial support. AA and HSOB thank the Oxford-Man Institute of Quantitative Finance for computing support.


\bibliography{example_paper}
\bibliographystyle{icml2023}

\newpage
\appendix
\onecolumn

\section{Additional figures}
\label{appendix:fig}

In this appendix we provide visualizations of the suggested Gromov-Hausdorff-informed graph search space.

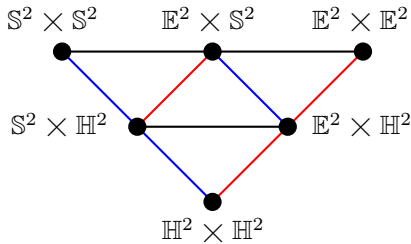
\begin{figure}[H]
    \centering
     \begin{tikzpicture}



            \draw[red, thick] (0,0) -- (1,1);
            \draw[red, thick] (1,1) -- (2,2);
            \draw[blue, thick] (0,0) -- (-1,1);
            \draw[blue, thick] (-1,1) -- (-2,2);
            \draw[black, thick] (-2,2) -- (0,2);
            \draw[black, thick] (0,2) -- (2,2);
            \draw[black, thick] (-1,1) -- (1,1);
            \draw[blue, thick] (0,2) -- (1,1);
            \draw[red, thick] (0,2) -- (-1,1);

            \filldraw[color=black, fill=black, very thick](0,0) circle (0.1);
            \draw (-8mm, -4mm) node[anchor=west] {$\mathbb{H}^2\bigtimes\mathbb{H}^2$};
            \filldraw[color=black, fill=black, very thick](1,1) circle (0.1);
            \draw (12mm, 10mm) node[anchor=west] {$\mathbb{E}^2\bigtimes\mathbb{H}^2$};
            \filldraw[color=black, fill=black, very thick](-1,1) circle (0.1);
            \draw (-28mm, 10mm) node[anchor=west] {$\mathbb{S}^2\bigtimes\mathbb{H}^2$};
        
            \filldraw[color=black, fill=black, very thick](0,2) circle (0.1);
            \draw (12mm, 24mm) node[anchor=west] 
            {$\mathbb{E}^2\bigtimes\mathbb{E}^2$};
            \filldraw[color=black, fill=black, very thick](2,2) circle (0.1);
            \draw (-8mm, 24mm) node[anchor=west] {$\mathbb{E}^2\bigtimes\mathbb{S}^2$};
            \filldraw[color=black, fill=black, very thick](-2,2) circle (0.1);
            \draw (-28.5mm, 24mm) node[anchor=west] 
            {$\mathbb{S}^2\bigtimes\mathbb{S}^2$};
    \end{tikzpicture}
    \label{fig:graphhh}
    \caption{Slice of the graph search space for latent geometries of dimension 4: product manifolds obtained using 2 models spaces of dimension 2. The graph edges are shown in different colours to depict a different degree of connectivity (black: $w_{\mathbb{E}^2,\mathbb{S}^2}$, red: $w_{\mathbb{E}^2, \mathbb{H}^2}$, blue: $w_{\mathbb{S}^2, \mathbb{H}^2}$), this is determined by the inverse of the Gromov-Hausdorff distance between the different product manifolds.}
\end{figure}

\begin{figure}[H]
    \centering\includegraphics[height=8.5cm,width=0.5\textwidth]{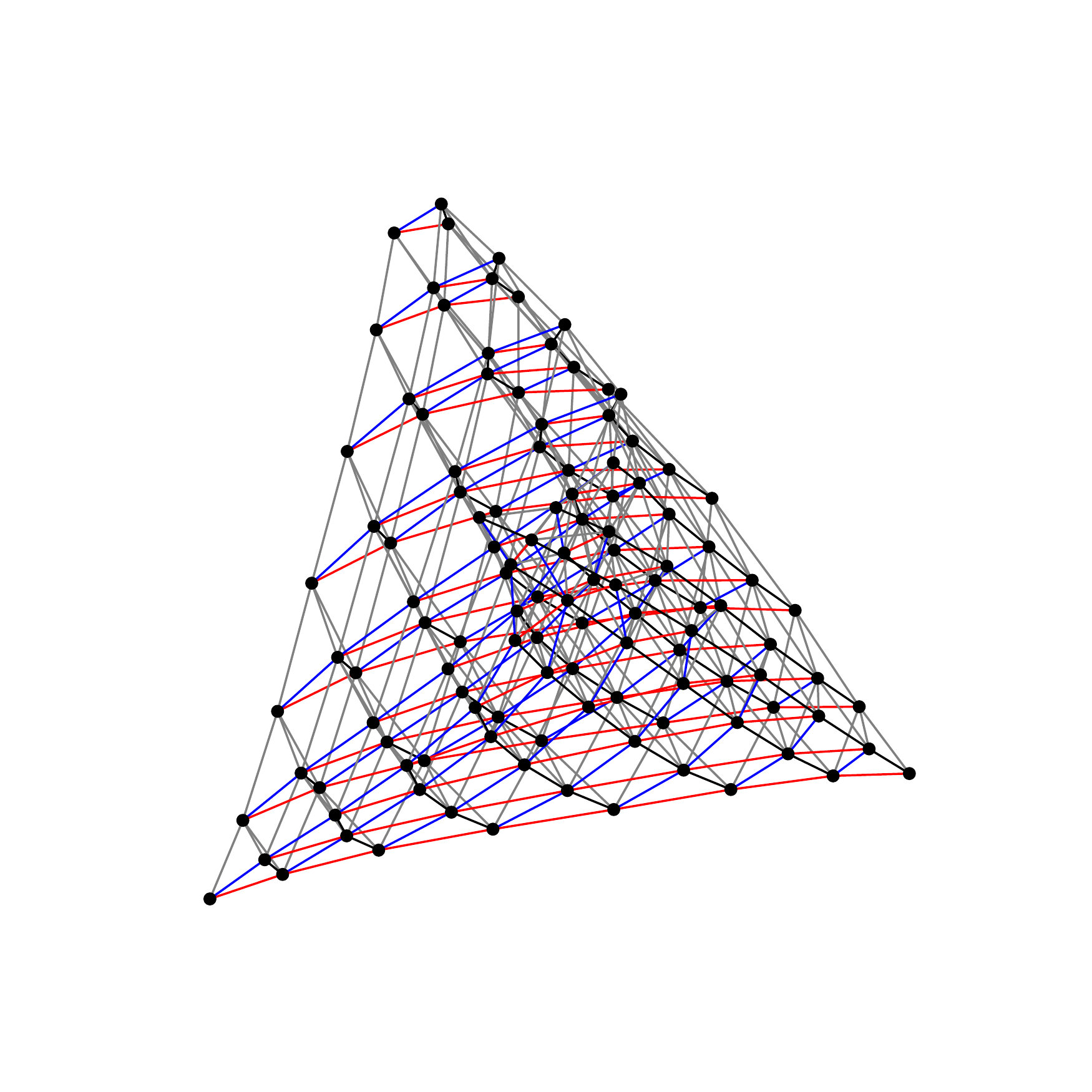}
    \vspace{-5mm}
   \caption{Example graph search space for product manifolds composed of up to seven model spaces. Manifolds of different dimensionality are connected with edges coloured grey. Node labels have been omitted for visual clarity. }
    \label{fig:example-search-space} 
\end{figure}

\section{Upper Bound for the Gromov-Hausdorff Distance}

\label{Upper Bound}

It is important to acknowledge that when considering both $A$ and $B$ as compact sets, there exists a straightforward upper bound for their Gromov-Hausdorff distance, which can be expressed in terms of their respective diameters. The \textit{diameter} of a metric space $Y$ is defined to be $\diam (Y)=\sup_{y, y'\in Y} d_Y(y, y')$. Given $a_0\in A$ and $b_0\in b$, we can define the isometric embeddings $f: A\rarrow A\times B$ and $g: B\rarrow A\times B$ given by $f(a)=(a, b_0)$ and $g(b)=(a_0, b)$. It is trivial that  $ \dh^{A\times B, f, g} (A, B)\leq \max\left(\diam (A), \diam (B)\right)$. Since  the  triple $(A\times B, f, g)$ belongs to  $\ES (A, B)$, we can estimate 
$\dhg (A, B)\leq\max\left(\diam (A), \diam (B)\right).$ 
In order to compare $\E^n$, $\H^n$, and $\S^n$, we suggest considering closed balls with a radius of one in each space. Since these balls, regardless of the space, are homogeneous Riemannian manifolds, they are isometric to one another. By estimating or establishing an upper bound for their Gromov-Hausdorff distance, we can effectively compare these spaces. With exactly an analogous argument as before, we can notice that given two compact balls of radius one $B$ and $B'$, of centres $x_0$ and $x_0'$, we can embed $B$ into $B\times B'$ by the mapping $f: b\mapsto (b, x_0')$. From here, it is obvious to see that $\dh^{B\times B', f, \id} (B, B\times B')=1$. Effectively, this observation provides us with the  bound $\dhg(B, B')\leq 1$. We will consider this as the estimation of Gromov-Hausdorff distances for product manifolds that differ solely in one model space, such as $\H^2$ and $\S^2\times \H^2$.

\end{document}